\documentclass[runningheads]{llncs}
\usepackage[T1]{fontenc}
\usepackage{graphicx}
\usepackage{longtable, makecell, multirow}
\usepackage{booktabs}
\usepackage{float}
\begin{document}
\title{Self-Supervised Learning for Pre-training Capsule Networks: Overcoming Medical Imaging Dataset Challenges}
\titlerunning{Self-Supervised Learning for Pre-training Capsule Networks}
%
\author{Heba El-Shimy\inst{1}\thanks{\email{he12@hw.ac.uk}} \and
Hind Zantout\inst{1} \and
Michael A. Lones\inst{2} \and
Neamat El Gayar\inst{1}}
%
%
\institute{Heriot-Watt University, Dubai, UAE \and
Heriot-Watt University, Edinburgh, Scotland, UK\\}
\maketitle              
\begin{abstract}
Deep learning techniques are increasingly being adopted in diagnostic medical imaging. However, the limited availability of high-quality, large-scale medical datasets presents a significant challenge, often necessitating the use of transfer learning approaches. This study investigates self-supervised learning methods for pre-training capsule networks in polyp diagnostics for colon cancer. We used the PICCOLO dataset, comprising 3,433 samples, which exemplifies typical challenges in medical datasets: small size, class imbalance, and distribution shifts between data splits. Capsule networks offer inherent interpretability due to their architecture and inter-layer information routing mechanism. However, their limited native implementation in mainstream deep learning frameworks and the lack of pre-trained versions pose a significant challenge. This is particularly true if aiming to train them on small medical datasets, where leveraging pre-trained weights as initial parameters would be beneficial. We explored two auxiliary self-supervised learning tasks---colourisation and contrastive learning---for capsule network pre-training. We compared self-supervised pre-trained models against alternative initialisation strategies. Our findings suggest that contrastive learning and in-painting techniques are suitable auxiliary tasks for self-supervised learning in the medical domain. These techniques helped guide the model to capture important visual features that are beneficial for the downstream task of polyp classification, increasing its accuracy by 5.26\% compared to other weight initialisation methods.

\end{abstract}
\section{Introduction}
The use of deep learning techniques in diagnostic medical imaging (DMI) is increasing, due to its superior performance compared to human physicians~\cite{khalifa-2024}. These models, which consist of billions of parameters, can capture minute details in images that may go unnoticed by the human eye. However, achieving such high levels of performance requires deep learning models to train on tens to hundreds of thousands of labelled images~\cite{shen-2017}. In the medical field, one common challenge to the wide adoption of deep learning models is the availability of such high-quality expert-annotated data, prompting the adoption of transfer learning approaches. Fine-tuning models that are pre-trained on a large, complex dataset usually offers a good initialisation for model parameters, facilitating learning from smaller datasets by subtly adjusting the layer weights to accommodate new data features. A frequently used dataset for pre-training is ImageNet~\cite{imagenet_cvpr09}, a visual concepts database comprising more than 14 million images with human-labelled annotations of objects in at least one million images. Although belonging to a different domain, ImageNet is widely used for pre-training deep learning models, often resulting in better weight initialisation compared to other strategies such as random, Xavier/Glorot~\cite{glorot-2010} or Kaiming/He~\cite{He2015DelvingDI} initialisations. Various studies have used ImageNet pre-trained models in DMI, achieving notable results irrespective of the downstream task, as evidenced by the work in~\cite{krenzer-2023}, among others. Many deep learning frameworks provide ImageNet pre-trained variants of popular architectures such as Convolutional Neural Networks (CNNs) or Visual Transformers (ViTs). However, attempting to leverage ImageNet pre-training advantages for DMI with less popular, or possibly entirely new architectures, poses a challenging, potentially costly task. For these models, pre-training on ImageNet is necessary prior to using them for the intended downstream task. 

This study investigates the use of self-supervised learning (SSL) for pre-training capsule neural network models (CapsNets). CapsNets are valued for their inherent interpretability owing to their architecture~\cite{lalonde-2020a,sabour-2017}. While other interpretable approaches exist, the unique ability of CapsNets to encode object instantiation parameters and spatial relationships is particularly valuable for DMI. However, CapsNets are under-represented as an architecture, lacking native implementations in most deep learning frameworks. Researchers utilising CapsNets often face the challenge of developing the architecture from scratch. Moreover, the quest for optimal weight initialisation through pre-training on a huge dataset like ImageNet is further complicated by CapsNets' computationally expensive training requirements. This can be an unfeasible pre-requisite, especially for researchers working with limited resources. Our contributions in this study include: (1) developing a novel CapsNet architecture for the classification of polyps in colon cancer; (2) developing two datasets of polyp frames that support different SSL auxiliary tasks, namely, colourisation and contrastive learning; and (3) evaluating the performance of our CapsNet architecture on polyp diagnosis under three experimental conditions: training from scratch, pre-training with SSL, and utilising initial layers from a ResNet pre-trained on ImageNet.

The structure of this paper is as follows: Section~\ref{sec2} reviews related work on CapsNets and SSL. Section~\ref{sec3} describes our methodology and experimental setup. Section~\ref{sec4} discusses our results. Section~\ref{sec5} provides conclusions and future work.

\section{Related Work}\label{sec2}

\subsection{Capsule Networks}
Capsule networks (CapsNets), as introduced by~\cite{sabour-2017}, are an advancement of CNNs, taking advantage of their powerful feature extraction capabilities by applying multiple convolutional filters to inputs. Recognising CNNs' pooling layer limitations in routing information between layers, CapsNets incorporate inverse graphics and dynamic routing to capture part-whole relationships and route information between layers. CapsNets replace individual neurons in CNNs with small groups of neurons that can encode various instantiation parameters such as position, size, orientation, albedo, hue, and texture. Primary capsules are tasked with learning these instantiation parameters for objects depicted in the inputs, effectively reversing the graphics rendering process, which typically uses such parameters for generating images from objects. Furthermore, the dynamic routing algorithm iteratively learns part-whole relationships by computing the similarity between each primary capsule and the higher-layer capsules. Capsules from lower layers that are parts of a larger whole will exhibit stronger connections to the corresponding higher-layer capsule representing that whole entity. Due to their use of inverse graphics and dynamic routing, CapsNets are considered inherently interpretable and capable of disentangling overlapping objects, as demonstrated in~\cite{sabour-2017}. In addition, they offer better explainability for their predictions, as discussed in~\cite{lalonde-2020a}.

CapsNets use margin loss, which ensures that a capsule of class $k$ is allowed to have a lengthy instantiation vector if and only if the entities associated with that class exist in the image. The total margin loss is the sum of all final layer capsule losses:
\begin{equation}
L_k = T_k \max(0, m^{+} - \|\mathbf{v}_k\|)^2 + \lambda (1 - T_k) \max(0, \|\mathbf{v}_k\| - m^{-})^2
\label{eq_margin_loss}
\end{equation}
where hyperparameters $T_k$, $m^+$, and $m^-$ are assigned the values 1, 0.9, and 0.1, respectively, in accordance with~\cite{sabour-2017}.

CapsNets may optionally incorporate a decoder that uses the output vector of the predicted class capsule to reconstruct the original input. The decoder uses mean squared error (MSE) loss, which, combined with the margin loss, form the total loss for the network. The decoder loss is usually weighted to prevent it from dominating the total loss. Additionally, it has demonstrated a regularisation effect that prevents the network from overfitting.

Several studies have successfully implemented CapsNets for DMI as summarised in~\cite{ElShimy2022ARO}. Research indicates that CapsNet surpass CNNs in computer-aided diagnostics~\cite{deepika-2022} and can be trained on small, imbalanced datasets~\cite{Afshar20203DMCNA3}. Furthermore, novel architectures have adapted CapsNet to address its limitations by improving scalability and enhancing efficiency~\cite{deepika-2022}.

\subsection{Self-Supervised Learning}
Self-supervised learning (SSL) is a subset of unsupervised learning where models are trained without labelled data. SSL employs auxiliary prediction (pre-text) tasks, which guide the model to autonomously generate task-relevant labels, effectively serving as the model's own supervision signal~\cite{krenzer-2023}. The quality of the features learnt is contingent upon the auxiliary task, with recent years witnessing several experiments refining these tasks. Among the predominant approaches is SimCLR, introduced by~\cite{chen-2020}. This technique presents the model with two pairs of the same image subjected to different augmentations. By employing contrastive loss the model learns to identify the matching pairs despite the augmentation:
\begin{equation}
\mathcal{L} = -\log \frac{\exp(sim(z_i, z_j)/\tau)}{\sum_{k \neq i}^{2N} \exp(sim(z_i, z_k)/\tau)}
\end{equation}
where $z_i$ and $z_j$ denote the encoded representations of the two augmented views of an image, $sim(z_i,z_j)$ refers to the cosine similarity between vectors $z_i$ and $z_j$, and $\tau$ is a temperature parameter. The values of $k \neq i$ exclude instances where $k=i$ from the summation to avoid comparing an embedding with itself. With $N$ representing the batch size, the expression $2N$ accounts for the doubled number of images per batch. SimCLRv2, introduced by~\cite{Chen2020BigSM} was pre-trained on ImageNet in a self-supervised manner, achieving remarkable results when fine-tuned with a few labelled examples. SimCLRv2 is commonly used in benchmarking techniques as in~\cite{caron-2021}.

Context prediction is another form of auxiliary tasks for SSL introduced in~\cite{Doersch2015UnsupervisedVR}. The model is given a pair of patches from the same image and is trained by learning to predict which of eight possible spatial relationships exist between the patches (e.g., south-east, west, north). In-painting is another auxiliary task described in~\cite{Pathak-2016}, where small patches of a few pixels are removed from images and the model learns to fill these patches by predicting the missing pixel values. Colourisation can also be used as a pretext task for SSL; the model receives greyscale images as input and learns to predict the pixel values corresponding to their coloured counterparts~\cite{Larsson2017ColorizationAA}.

SSL has proven to be an effective method for pre-training models, achieving top-1 accuracy on ImageNet (predicting the correct class with the highest confidence). Using SSL for pre-training allows models to learn rich, generalisable, and robust representations from the data, capturing important visual features. In some instances, emerging properties such as semantic segmentation of images could be learnt with SSL, as demonstrated in~\cite{caron-2021}. SSL pre-training serves as an initial step to fine-tune models for downstream tasks, employing few-shot (1-5 samples) and low-shot (10-100 samples) learning techniques~\cite{krenzer-2023}. This method could be particularly effective in the medical domain, where data availability is constrained. 

\section{Self-Supervised Learning as a Pre-training Technique for Capsule Networks}\label{sec3}

\subsection{Dataset}
We used the PICCOLO dataset, which consists of 3,433 frames of 74 distinct polyps collected from 40 different patients at the Hospital Universitario Basurto in Bilbao, Spain~\cite{sanchez-peralta-2020}. The dataset includes high-resolution frames captured in white light (WL) and narrow band imaging (NBI) modalities. Associated with the data set is a metadata file that contains detailed annotations for each polyp, including its initial diagnosis by a gastroenterologist, the confirmed diagnosis from a pathologist, the polyp size, classifications according to the NICE~\cite{nice-2014} and Paris~\cite{paris-update} schemes, and the histological stratification. The dataset is pre-divided into a training set with 2,203 frames, a validation set with 898 frames, and a test set with 333 frames. A notable problem with this dataset is its small size, which poses challenges to deep learning models to accurately capture complex polyp features. In addition, the dataset presents three diagnosis classes with high intraclass variability and low interclass variability, adding to the complexity. There is a substantial imbalance within the dataset: the NICE Type 2 (adenoma) class constitutes over 72\% of the training data, while NICE Type 1 (hyperplasia) and NICE Type 3 (adenocarcinoma) make up 20\% and 8\%, respectively. The validation set exhibits a different distribution, but the dominant class remains NICE Type 1 at 66\%, while Types 1 and 3 make up 15\% and 19\%, respectively---contrasting with the training set. The challenge of class imbalance is further compounded by the shift in the data distribution of the test set. Type 3 becomes the dominant class in the test set, at 38\%, followed by Type 1 at 34\%, and Type 2, previously dominant in training and validation, now the minority class at merely 28\%.

\subsection{Dataset Pre-processing and Augmentation}
We developed a pre-processing pipeline that was uniformly applied across all experiments. Initially, we removed non-informative black borders, accounting for 15-20\% of each frame's area. Any residual black pixels at the edges were replaced with the average value of all non-black pixels within the frame. Next, we resized all frames to $224\times224$ pixels. Then, a mild Gaussian blur was applied for noise reduction, followed by contrast-limited adaptive histogram equalization (CLAHE)~\cite{pisano-1998} which enhanced contrast and texture to improve the model's feature detection capability. Given that the frames include a mix of WL and NBI modalities with differing histogram characteristics, and considering the strength of NBI in highlighting critical visual features of pre-cancerous and cancerous polyps, we developed an algorithm to automatically identify WL images and perform histogram matching using the closest NBI frame of the same polyp, ensuring dataset consistency. Finally, pixel values were rescaled to values between $0$ and $1$.

We applied a set of random augmentations solely on the training set. These augmentations encompassed: 1) random cropping which retains the image size at $224\times224$ and ensures no less than 75\% of a frame is cropped, thus preserving the visual features and context of polyps; 2) random rotation up to ($\pm25^\circ$); 3) random affine transformations with a maximum of $3^\circ$ and a shear value of 7; 4) random perspective change with a distortion scale of 0.15 and probability of 0.25; 5) random vertical and horizontal flips each with a probability of 0.35; and 6) random colour jitter where only the brightness, contrast, and saturation are modified by a value of 0.2, while the hue remains unchanged to avoid the loss of crucial polyp features.

\subsection{Model Architecture}
The original CapsNet architecture proposed in~\cite{sabour-2017} represents a simple architecture intended to recognise handwritten digits from low-resolution images of $28\times28$ pixels. The hyperparameters were specifically tailored for this task; for instance, the kernel size was relatively high and the stride was large. The decoder consisted of three fully-connected layers of relatively small number of neurons. However, when applied to higher-resolution images, these hyperparameters tend to skip the finer details during feature extraction. Additionally, using the original architecture with high-resolution images becomes computationally expensive because the number of primary capsules increases quadratically, influenced by the spatial dimensions of the feature maps.

\begin{figure}[h]
    \centering
    \includegraphics[width=1\linewidth]{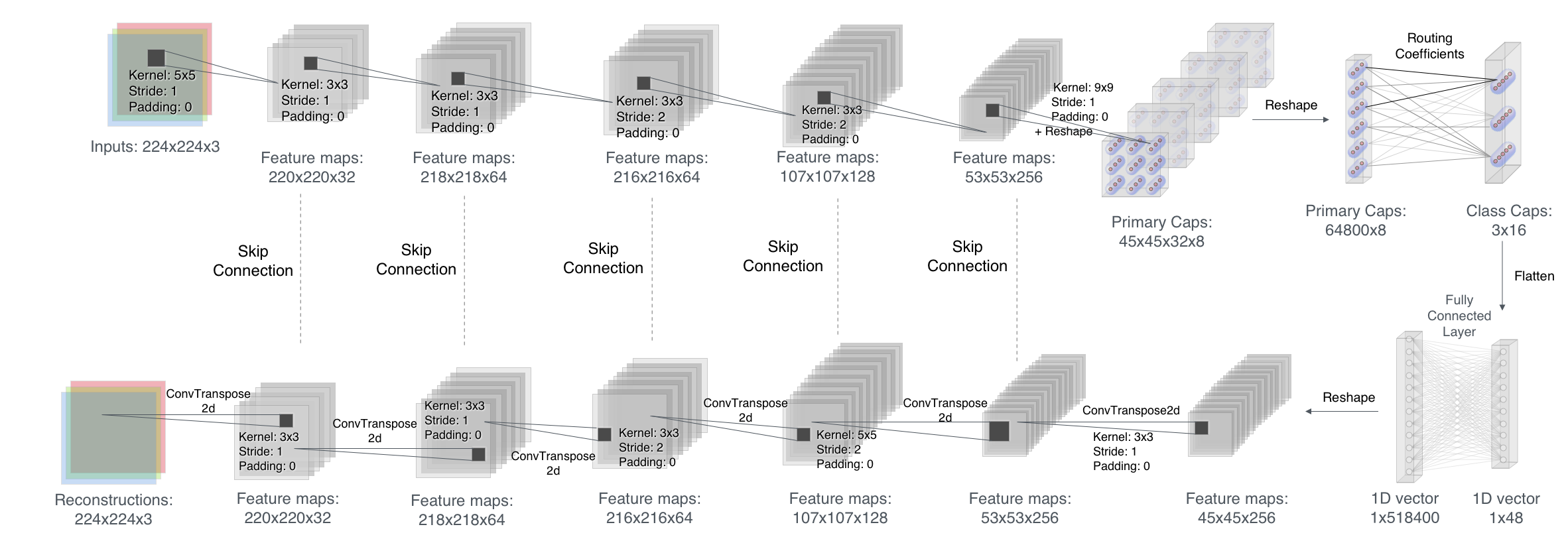}
    \caption{Modified capsule network}
    \label{fig:ModCapsNet}
\end{figure}

We developed a modified CapsNet architecture, shown in Fig.~\ref{fig:ModCapsNet}. This architecture incorporates five additional convolutional layers before the primary capsules layer, using smaller kernel sizes and strides. This setup enables the model to capture the fine details of polyps. In the decoder, we used several transposed convolutions instead of fully connected layers in the original architecture. We added skip connections between the encoder and decoder, but down-weighted their contribution to the decoder feature maps. This approach helps the network improve its reconstruction capability instead of simply replicating features from the encoder.

In our classification experiments, we substituted the margin loss in~\cite{sabour-2017} with the spread loss in a subsequent CapsNet architecture presented in~\cite{hinton-2018}:
\begin{equation}
L_i = (\max(0, m - (a_t - a_i)))^2, \quad L = \sum_{i\neq t} L_i
\end{equation}
where $m$ is the margin, $a_t$ is the activation for the target class, and $a_i$ is the activation of a wrong class.
Spread loss aims to increase the gap between the correct class and the incorrect ones while incorporating an adaptive margin mechanism. Spread loss was found to be more suitable for multi-class classification tasks with more stable training and less sensitivity to weight initialisation and hyperparameter values.

\subsection{SSL Pretext Tasks}
We decided to investigate two SSL auxiliary tasks that are suitable for the downstream task of polyp classification. These included experiments on colourisation and contrastive learning as in SimCLR~\cite{chen-2020} combined with an in-painting task. Polyps are usually categorised on the basis of visual characteristics such as colour, surface patterns, and texture. The choice of colourisation and contrastive learning with in-painting tasks was based on the intuition that these tasks could guide the model to learn the colour and texture of polyps in images, capturing useful information that would support the downstream task.

\paragraph{Colourisation}:
For this task, the model is fed a batch of greyscale images along with their corresponding coloured versions as ground truth. The objective of the model is to learn the mapping of pixel values required to produce the coloured version. We used a staged-learning approach that incorporates multiple losses throughout the training inspired by~\cite{wang-2019}. Two loss functions were selected to form the reconstruction loss: L1 loss and perceptual loss. L1 loss, or mean absolute error (MAE) was chosen over MSE used in~\cite{sabour-2017} due to its lower sensitivity to outliers. Perceptual loss computes the MSE between feature maps extracted from different layers of a small, pre-trained CNN (VGG16) for both the reconstructed images and the ground truth.

Based on promising results from initial experiments, training was carried out in two stages. For the first 30\% of epochs, we gradually introduced perceptual loss by increasing its weight with every epoch:
\begin{equation}
\lambda_{p} = \min(0.1, \frac{epoch}{30} \times 0.1)
\end{equation}
\begin{equation}
\mathcal{L}_{reconstruction} = \lambda_{p}\mathcal{L}_{perceptual} + \mathcal{L}_{1}
\end{equation}
For the remaining epochs, the losses are weighted as follows:
\begin{equation}
\mathcal{L}_{reconstruction} = 0.1 \mathcal{L}_{perceptual} + \mathcal{L}_{1}
\end{equation}

Through this colourisation task, the model learns important visual features for polyp classification, such as vessel patterns and colour characteristics~\cite{nice-2014}.

\paragraph{Contrastive learning with in-painting}:
The model processes batches containing pairs of images subjected to distinct augmentations and masked/black patches. The objective is learning to identify the corresponding image pairs within each batch through a contrastive loss function, specifically the normalised temperature-scaled cross entropy (NT-Xent) proposed by~\cite{chen-2020}. This loss function operates on image feature embeddings, minimising the distance between semantically similar images while simultaneously maximising the distance between dissimilar ones as below:

\begin{equation}
\ell_{i,j} = -\log\frac{\exp({z_i^T z_j/\tau})}{\sum_{k=1}^{2N}1_{[k\neq i]}\exp({z_i^T z_k/\tau})}
\end{equation}
where the cosine similarity between two normalised embeddings $z_i$ and $z_j$ is defined as: 
\begin{equation}
z_iT z_j = \frac{z_i^T z_j}{||z_i|| \cdot ||z_j||}
\end{equation}
The aggregate contrastive loss for all positive pairs in a batch is computed as:
\begin{equation}
\mathcal{L}_{contrastive} = \frac{1}{2N}\sum_{k=1}^N[\ell_{2k-1,2k} + \ell_{2k,2k-1}]
\end{equation}
To enhance the capability of the model of in-painting the missing regions, we incorporate the staged reconstruction loss approach used in the colourisation task. The final loss function combines both losses:
\begin{equation}
\mathcal{L}_{total} = \mathcal{L}_{contrastive} + \mathcal{L}_{reconstruction}
\end{equation}

This auxiliary task encourages the model to learn visual features such as texture, colour, and surface patterns, which are crucial for polyp classification~\cite{nice-2014}.

\subsection{Experimental Setup}
We used the AdamW optimiser~\cite{Loshchilov2017DecoupledWD} in conjunction with a cosine annealing learning rate scheduler with warm restarts~\cite{Loshchilov2016SGDRSG}. The learning rate follows a cyclical pattern, initialising at $10^{-3}$ and gradually decreasing to a minimum of $10^{-6}$ before resetting, with the period of each subsequent cycle doubling. Due to computational constraints, we utilised a batch size of 8, which is suboptimal for CapsNets to learn effectively. To address this limitation, we implemented gradient accumulation over 8 consecutive batches, effectively simulating a batch size of 64. To mitigate class imbalance, we developed a custom weighted sampler that ensures uniform class distribution within each batch during the training phase. The SSL models were trained for 200 epochs, while the supervised classification models were trained for a minimum of 50 epochs. We implemented an early stopping strategy for the supervised models based on the validation loss to prevent overfitting. All experiments were conducted on dual NVIDIA A6000 GPUs with 48GB of VRAM each. Hyperparameters were selected through extensive and systematic experimentation, optimising model performance on a validation subset of the PICCOLO dataset.

\section{Results and Discussion}\label{sec4}
Our CapsNet architecture was trained for the supervised classification of polyps using four weight initialisation strategies: 1) Kaiming/He initialisation for the convolutional layers with ReLU activation combined with Glorot/Xavier initialisation for both the primary capsule layer and routing weights; 2) ImageNet pre-trained ResNet weights applied to the first four convolutional layers only; 3) SSL colourise pre-trained weights applied to all layers; and 4) SSL contrastive pre-trained weights applied to all layers.

\begin{figure}[ht]
    \centering
    \includegraphics[width=0.45\linewidth]{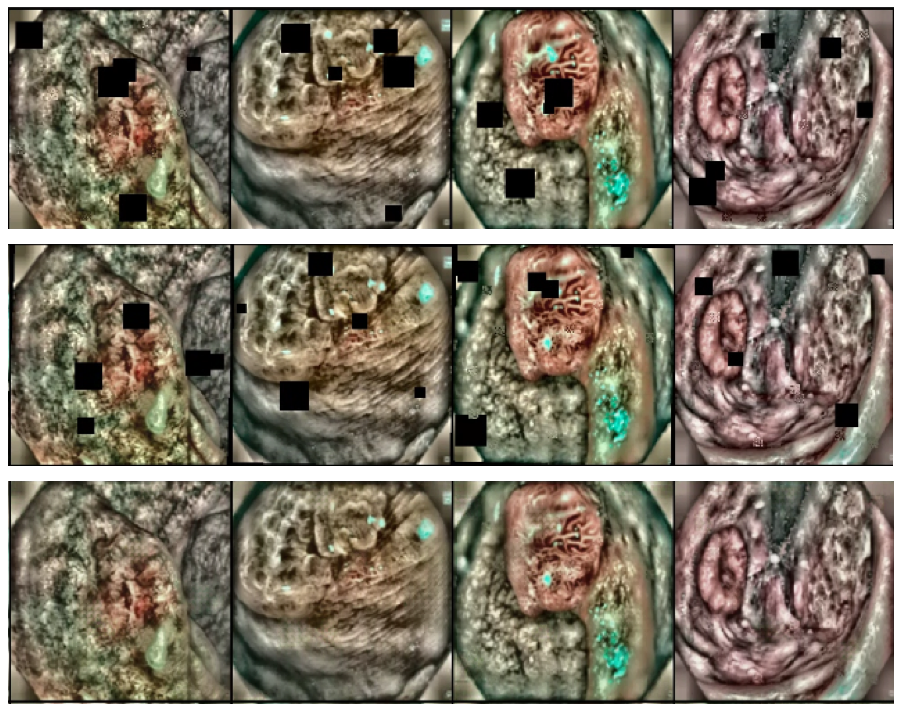}
    \caption{Outputs from our modified CapsNet architecture trained by combining SSL contrastive and in-painting auxiliary tasks. The top and middle rows show pairs of differently augmented versions of the same images. The bottom row demonstrates the network's in-painting capability after 200 epochs of training.}
    \label{fig:contrastive}
\end{figure}

\begin{figure}[!h]
    \centering
    \includegraphics[width=0.45\linewidth]{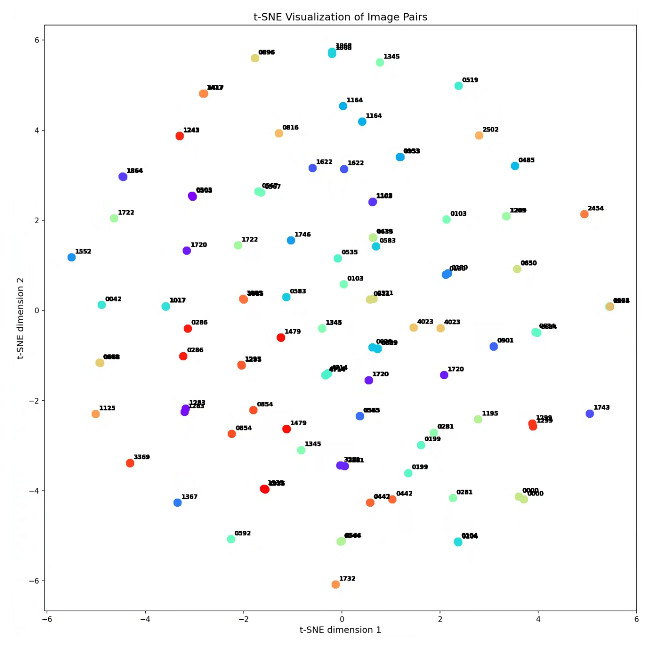}
    \caption{t-SNE visualisation of feature embeddings from our modified CapsNet model after SSL contrastive learning and in-painting auxiliary task. Each dot represents an image projected into 2D space, with matching numeric IDs and colours indicating pairs of the same image.}
    \label{fig:contrastive_plot}
\end{figure}

As illustrated in Fig.~\ref{fig:contrastive}, CapsNet demonstrated enhanced feature extraction capabilities when trained using SSL contrastive learning combined with the in-painting auxiliary task. While the network successfully reconstructed most missing patches, it was unable to fully capture the full vibrancy of the colours in the original image, suggesting that extended training time may be beneficial. Furthermore, CapsNet effectively learned to identify corresponding image pairs, as evidenced by the reduced distances between the matching data points in the manifold space shown in Fig.~\ref{fig:contrastive_plot}.

\begin{figure}[!h]
    \centering
    \includegraphics[width=0.45\linewidth]{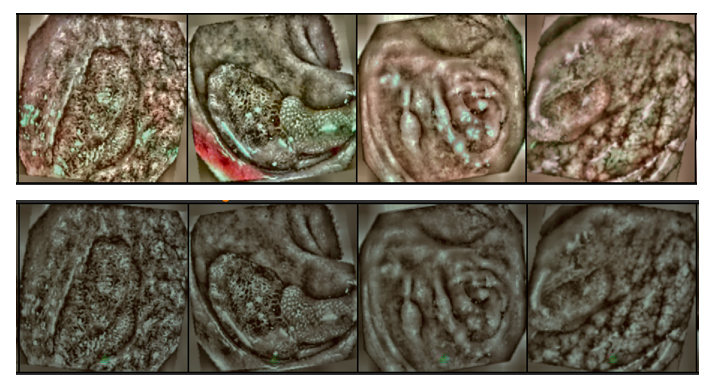}
    \caption{Outputs of our modified CapsNet after 200 epochs of training on SSL colourise task}
    \label{fig:colourise}
\end{figure}

In our experiments with the colourisation auxiliary task, the network successfully captured and reconstructed structural features but struggled to accurately colourise the greyscale images, as shown in Fig.~\ref{fig:colourise}. This suggests that contrastive learning and in-painting are more effective as SSL auxiliary tasks. The difficulty likely stems from the complexity of simultaneously modifying pixel values across all three colour channels (red, green, blue) for relatively high-resolution $224\times224$ images. 

To evaluate the performance of our models, we used multiple metrics including accuracy, macro-averaged F1 score, specificity, and area under the receiver operating characteristic curve (AUROC). However, due to dataset imbalance, we primarily focus on Matthew's correlation coefficient (MCC) and balanced accuracy as the key evaluation metrics. Table~\ref{tab:test-performance} summarises the performance of the models on the test set.

\begin{table}
  \centering
  \caption{Performance comparison of CapsNets using different initialisation strategies, evaluated on the PICCOLO test set.}
  \begin{tabular}{>{\raggedright\arraybackslash}m{1.85cm} >{\raggedright\arraybackslash}m{1.4cm} >{\raggedright\arraybackslash}m{1.3cm} >{\raggedright\arraybackslash}m{1.8cm} >{\raggedright\arraybackslash}m{1.5cm} >{\raggedright\arraybackslash}m{1.5cm} >{\raggedright\arraybackslash}m{1.8cm}}
    \toprule
    Model & Accuracy & MCC & Balanced Accuracy & AUROC & F1-score (Macro Avg.) & Specificity \\
    \midrule
    Kaiming+ & 0.38 & 0.16 & 0.12 & 0.59 & 0.34 & 0.69 \\
    Xavier & & & & & & \\
    \addlinespace[0.6em]
    ImageNet & 0.38 & 0.25 & 0.15 & 0.71 & 0.33 & 0.69 \\
    \addlinespace[0.6em]
    SSL contrastive pre-trained & 0.40 & 0.22 & 0.16 & 0.62 & 0.34 & 0.70 \\
    \addlinespace[0.6em]
    SSL colourisation pre-trained & 0.38 & 0.00 & 0.00 & 0.52 & 0.18 & 0.69 \\
    \bottomrule
  \end{tabular}
  \label{tab:test-performance}
\end{table}

From the results, the ImageNet-pretrained and SSL-contrastive pre-trained models perform comparably. The SSL-contrastive pre-trained model has higher accuracy (0.40 vs 0.38), fractionally higher balanced accuracy (0.16 vs 0.15), and marginally higher specificity (0.70 vs 0.69) indicating good detection of negative samples. The ImageNet-pretrained model achieved higher MCC (0.25 vs 0.22) indicating stronger correlation between predictions and labels, and notably higher AUROC (0.71 vs 0.62) suggesting better classification capability. The SSL-colourisation pre-trained model is the worst performing, with results approaching random prediction (AUROC: 0.52, MCC: 0.00). Notably, ImageNet pre-trained models' performance on PICCOLO is not necessarily much higher than training from scratch. The comparable performance between SSL-contrastive and ImageNet-pretrained models is promising, considering that SSL pre-training was performed on a significantly smaller dataset than ImageNet, suggesting this technique warrants further exploration.

During our investigation of CapsNet training dynamics, we observed significant sensitivity to parameter initialisation and slower convergence rates compared to traditional architectures. Monitoring weight updates revealed minimal changes in both primary capsules and routing weights across consecutive epochs, with variations typically ranging from $10^{-6}$ to $10^{-5}$. This effect was particularly pronounced in deeper architectures, where the gradient signal weakened considerably before reaching the class capsules.

The challenge was further exacerbated during SSL, where the absence of direct classification loss signals resulted in a reduced engagement of class capsules during training. While class capsules exhibited responses to the rest of the network's learning of auxiliary tasks, these adaptations were notably subtle. This could be addressed by extending the SSL training time for CapsNets from 200 to 500 epochs.

Further analysis revealed that class capsules implicitly learnt classification-relevant features during SSL training, as evidenced by network accuracy and MCC tracking. However, we observed that the capsules also learnt the dataset imbalances, which proved detrimental to subsequent training with transfer learning. When using weights pre-trained on imbalanced data, the network required an additional 5 to 10 epochs to overcome these inherent biases. This performance degradation was more severe compared to other initialisation strategies.

\section{Conclusions}\label{sec5}
Our investigation of SSL for pre-training CapsNet, particularly for DMI applications, has revealed several important considerations for effective pre-training. The results indicate that CapsNets require careful optimisation strategies for SSL auxiliary tasks, due to their unique training dynamics. However, our findings demonstrate that CapsNets can slowly and gradually learn important visual features when trained using a suitable SSL auxiliary task, particularly through contrastive learning combined with an in-painting task. Our SSL-contrastive pre-training approach, despite using a significantly smaller dataset, achieved comparable results to ImageNet pre-training, even surpassing it with increases of 5.26\% in accuracy, 6.67\% in balanced accuracy and 1.45\% in specificity. The results indicate that SSL pre-training is a viable technique that could be investigated further.

Future work will explore longer training periods and implement layer-specific learning rate adjustments to better accommodate the slow learning nature of CapsNets. We plan to investigate the impact of dataset balancing during SSL pre-training to address the challenges posed by implicit learning of data imbalances. The proposed experiments aim to establish more effective SSL strategies for CapsNet, which may open up new opportunities for transfer learning in DMI applications. Ultimately, this research contributes towards improving the performance of deep learning models on limited and challenging medical datasets.

%
%
%
\nocite{*}
\bibliographystyle{splncs04}
\bibliography{ssl}
\end{document}